\documentclass[11pt,a4paper,AutoFakeBold,UTF8]{article}
\usepackage[hyperref]{acl2021}
\usepackage{times}
\usepackage{latexsym}

\usepackage{microtype}

\usepackage{graphicx}
\usepackage{float}
\usepackage{amsmath}
\usepackage{amssymb}
\usepackage{multirow}
\usepackage{booktabs}
\usepackage{url}
\usepackage{array}
\usepackage{enumitem}
\usepackage{algorithm}
\usepackage{algorithmic}
\usepackage{pifont}
\usepackage{CJKutf8}
\usepackage{CJK,CJKnumb} 
\usepackage[normalem]{ulem}
\usepackage{makecell}
\usepackage{stackengine}
\usepackage{bm}
\usepackage{subfigure}

\aclfinalcopy 

\title{ExpMRC: Explainability Evaluation for Machine Reading Comprehension}

\author{Yiming Cui$^{1,2}$, Ting Liu$^1$, Wanxiang Che$^1$, Zhigang Chen$^{2}$, Shijin Wang$^{2,3}$ \\ 
{$^1$Research Center for SCIR, Harbin Institute of Technology, Harbin, China} \\
{$^2$State Key Laboratory of Cognitive Intelligence, iFLYTEK Research, China} \\
{$^3$iFLYTEK AI Research (Hebei), Langfang, China} \\
{$^1$\tt\{ymcui,tliu,car\}@ir.hit.edu.cn} \\
{$^{2,3}$\tt\{ymcui,zgchen,sjwang3\}@iflytek.com}\\  
}

\date{}

\begin{document}
\begin{CJK*}{UTF8}{gkai}
\maketitle
\begin{abstract}
Achieving human-level performance on some of Machine Reading Comprehension (MRC) datasets is no longer challenging with the help of powerful Pre-trained Language Models (PLMs).
However, it is necessary to provide both answer prediction and its explanation to further improve the MRC system's reliability, especially for real-life applications.
In this paper, we propose a new benchmark called ExpMRC for evaluating the explainability of the MRC systems.
ExpMRC contains four subsets, including SQuAD, CMRC 2018, RACE$^+$, and C$^3$ with additional annotations of the answer's evidence.
The MRC systems are required to give not only the correct answer but also its explanation.
We use state-of-the-art pre-trained language models to build baseline systems and adopt various unsupervised approaches to extract evidence without a human-annotated training set.
The experimental results show that these models are still far from human performance, suggesting that the ExpMRC is challenging.\footnote{Resources will be available through \url{https://github.com/ymcui/expmrc}} 
\end{abstract}

\section{Introduction}
Machine Reading Comprehension is a task that requires the machine to read and comprehend given passages and answer the questions, which receives wide attention over the past few years.
We have seen tremendous efforts on creating challenging datasets \citep{hermann-etal-2015,hill-etal-2015,rajpurkar-etal-2016,lai-etal-2017,cui-emnlp2019-cmrc2018,sun-etal-2020-c3} and designing effective models \citep{kadlec-etal-2016,cui-acl2017-aoa,seo-etal-2016}.

However, though the state-of-the-art systems could achieve better performance than the average human on some MRC datasets with the help of pre-trained language models \citep{devlin-etal-2019-bert,liu2019roberta,clark2020electra}, the explainability of these systems remains uncertain, which raises concerns in utilizing these models in real-world applications.
In a realistic view, question answering (QA) or MRC systems that only give final prediction could not convince the users since these results lack explainability. 
In this context, Explainable Artificial Intelligence (XAI) \citep{gunning2017explainable} received much more attention in these years than ever.
XAI aims to produce more explainable machine learning models while preserving a high accuracy of the model output and let the human understand its intrinsic mechanism.

Understanding the intrinsic mechanism of the neural network is a challenging issue.
There are several hot discussions on the relevant topics, such as {\em whether the attention could be explanations} \citep{serrano-smith-2019-attention,jain-wallace-2019-attention,wiegreffe-pinter-2019-attention,bastings-filippova-2020-elephant}.
Nonetheless, we could seek post-hoc explainability approaches, which target models that are not readily interpretable by design.
Post-hoc approaches resort to diverse means to enhance the model's interpretability \citep{barredo-etal-2020}.
One of the suitable post-hoc approaches for NLP is to generate text explanations, which is a practical way to alleviate the absence of the neural network's explainability. 
Though the text explanation does not necessarily interpret the model's intrinsic mechanism, it is informative to know both the predicted answer and its text explanation, especially for those real-life applications.

To better evaluate the MRC model's explainability, in this paper, we propose a comprehensive benchmark ExpMRC for the machine reading comprehension in a multi-lingual and multi-task way, which evaluates the accuracy of both answer and its explanation.
The proposed ExpMRC contains four subsets, including SQuAD \citep{rajpurkar-etal-2016}, CMRC 2018 \citep{cui-emnlp2019-cmrc2018}, RACE$^+$, and C$^3$ \citep{sun-etal-2020-c3} with additional annotations of the evidence spans, covering span-extraction MRC and multi-choice MRC in both English and Chinese. 
The MRC model should not only give an answer span or select a choice for the question but also give a passage span as the evidence, which creates more challenges to the machine.
The resulting dataset contains 11K human-annotated evidence spans over 4K questions.
The contributions of our paper are as follows.
\begin{itemize}
	\item We propose a new MRC benchmark called ExpMRC, which aims to evaluate the accuracy of the final answer prediction and how well it gives for its explanation.
	\item We also propose several baseline systems that adopt unsupervised approaches for ExpMRC.
	\item The experimental results on ExpMRC show that the current pre-trained language models are still far from satisfactory in giving explanations for the predicted answer, suggesting that the proposed ExpMRC is challenging.
\end{itemize}

\section{Related Work}
Machine reading comprehension has been regarded as an important task to test how well the machine comprehends human languages.
In the earlier stage, as most of the models \citep{dhingra-etal-2017,kadlec-etal-2016,cui-acl2017-aoa} are solely trained on the training data of each dataset without much prior knowledge, their performances are not very impressive.
However, as the pre-trained language models emerged during these years, such as BERT \citep{devlin-etal-2019-bert}, RoBERTa \citep{liu2019roberta}, ELECTRA \citep{clark2020electra}, etc., many systems could achieve better performances than average human on several MRC datasets, such as SQuAD 1.1 and 2.0 \citep{rajpurkar-etal-2016,rajpurkar-etal-2018-know}.

After reaching the `over-human' performance, there is another issue to be addressed. 
The decision process and the explanation of these artifacts still remain unclear, raising concerns about their reliability. 
In this context, XAI becomes more important than ever not only in NLP but also in various directions in artificial intelligence.
However, most of the cutting-edge systems are developed on neural networks, and the explainability investigation of these approaches is non-trivial, which is still an on-going research topic.

In NLP, some researchers conducted analyses to better understand the internal mechanism of BERT-based architecture.
For example, \citet{kovaleva-etal-2019-revealing} discovered that there are repetitive attention patterns across different heads in the multi-head attention mechanism indicating its over-parametrization. 
But perhaps the hottest discussion is {\em whether the attention could be explanations}. 
Some researchers argue that the attention could not be used as explanations, such as \citet{jain-wallace-2019-attention} verify that using completely different attention weights could also achieve the same prediction.
On the contrary, some works hold positive attitudes about this topic \citep{wiegreffe-pinter-2019-attention,bastings-filippova-2020-elephant}.
These works have brought us different views of attention-based models, but there is still no consensus about this important topic.

In MRC, the most relevant effort in explainability is the creation of HotpotQA \citep{yang-etal-2018-hotpotqa}, which is a multi-hop explainable question answering dataset.
HotpotQA requires the machine to retrieve the relevant documents and extract a passage span as the answer along with its evidence sentences.
Various models \citep{qiu-etal-2019-dynamically,shao-2020-gnn-analysis} have been proposed to tackle this task using supervised learning approaches with the labeled training data. 
However, unfortunately, almost all works focus on achieving higher scores in the benchmark without specifically caring about the explainability.

Above all, though various efforts have been made, we argue that the explainability is a universal demand for all machine reading comprehension tasks and different languages but not only restricted to English multi-hop QA. 
Another issue is that annotating evidence for each task is not feasible, and we should also seek unsupervised approaches, which do not rely on any annotated evidence to minimize the labor cost.

In this context, we propose ExpMRC to specifically focus on evaluating explainability on four tasks, covering span-extraction and multi-choice MRC in both English and Chinese.
ExpMRC does not provide annotated training data, which aims to enforce our community to focus on designing unsupervised approaches to improve their generalizability across different MRC tasks and even different languages.
To the best of our knowledge, this is the first MRC benchmark in a multi-task and multi-lingual setting, which could be used in not only explainability evaluation but also various other directions, such as cross-lingual studies, etc.

\begin{table*}[ht]
\small
\begin{center}
\begin{tabular}{p{0.08\textwidth}<{\centering} p{0.56\textwidth} l}
\toprule
\bf Subset & \bf Passage & \bf Question \& Answer \\
\midrule
\multirow{6}{*}{\bf SQuAD} & ... Competition amongst employers tends to drive up wages due to the nature of the job, since there is a relative shortage of workers for the particular position. {\bf \textcolor{blue}{Professional and labor organizations} may limit the supply of workers which results in higher demand and greater incomes for members.} Members may also receive higher wages through collective bargaining, political influence, or corruption ...
 	& \makecell[lt]{{\bf Q}: Who works to get workers \\ higher compensation? \\ {\bf A}: Professional and labor \\organizations}  \\
\midrule
\multirow{5}{*}{\makecell[ct]{\bf CMRC \\ \bf2018}} & ... 钩盲蛇（学名：``Ramphotyphlops braminus"）是蛇亚目盲蛇科下的一种无毒蛇种，主要分布在非洲及亚洲，不过现在钩盲蛇的分布已推广至世界各地。{\bfseries 钩盲蛇是栖息于\textcolor{blue}{地洞}的蛇种}，由于体型细小，加上善于掘洞，因此经常被误认为蚯蚓，唯一分别就是钩盲蛇的身体并没有分成明显的段节。...  
	& \makecell[lt]{{\bf Q}: 钩盲蛇一般生活在什么\\ 地形中？ \\ {\bf A}: 地洞} \\
\midrule
\multirow{9}{*}{\bf RACE$^+$} & ... By making their leaves, flowers, roots and fruits poisonous to enemies, plants can fight back. One such plant is the Golden Wattle tree, British scientist David Caneron has found when an animal eats the tree's leaves, the amount of poison increase in the other leaves. “It's like the injured leaves telephoning the others telling them to fight together against the enemy,” he said. {\bf The tree also sends defense messages to neighboring plants by giving out a special smell.} Golden Wattle trees in the nearby 45 meters will get the message and produce more poison within 10 minutes ...  
	& \makecell[lt]{{\bf Q}: According to the study, if one \\ Golden Wattle tree is attacked  \\ by animals, it can? \\ {\bf A}: tell other trees to protect it \\ {\bf B}: produce more poison within \\10 minutes \\ \bf C: sent defense messages to the \\ \bf neighboring plants \\ {\bf D}: kill the animals with its leaves} \\
\midrule
\multirow{7}{*}{\bf C$^3$} & ... 大学生活是走上社会的预演，可以说，{\bfseries 大学里的处世态度和人际关系的成功与否，直接决定着将来在社会上的成败。}人是社会性的动物，生活中的每个人都离不开别人的帮助，同时也在帮助着别人。不管是学习、生活、工作，都要求自己要有良好的处理人际关系的能力。一个人要想有良好的人际关系，就要遵循以下几个原则：一是“主动”。要主动和别人交往，主动帮助别人。二是“诚信”。 ...  
	& \makecell[lt]{{\bf Q}: 说话人认为什么因素决定\\ 在社会上的成败? \\ {\bf A}: 工作的态度\\ {\bf B}: 朋友的数量\\ {\bf C}: 大学里的学习成绩\\ \bf D: 大学里的人际关系} \\
\bottomrule
\end{tabular}
\caption{\label{expmrc-exmaple} Examples in ExpMRC. The evidence of the answer (in passage) and correct option (for multi-choice MRC) are marked in boldface, and answer spans (for span-extraction MRC) are marked in blue. }
\end{center}
\end{table*}

\begin{table*}[htp]
\small
\begin{center}
\begin{tabular}{l cccccccc}
\toprule
\multirow{2}*{\bf } & \multicolumn{2}{c}{\bf SQuAD} & \multicolumn{2}{c}{\bf CMRC 2018}& \multicolumn{2}{c}{\bf RACE$^+$} & \multicolumn{2}{c}{\bf C$^3$} \\
& \bf Dev & \bf Test & \bf Dev & \bf Test & \bf Dev & \bf Test & \bf Dev & \bf Test \\
\midrule
Language & \multicolumn{2}{c}{English} & \multicolumn{2}{c}{Chinese} & \multicolumn{2}{c}{English} & \multicolumn{2}{c}{Chinese} \\
Answer Type & \multicolumn{2}{c}{passage span} & \multicolumn{2}{c}{passage span} & \multicolumn{2}{c}{multi-choice} & \multicolumn{2}{c}{multi-choice} \\
Domain & \multicolumn{2}{c}{Wikipedia} & \multicolumn{2}{c}{Wikipedia} & \multicolumn{2}{c}{exams} & \multicolumn{2}{c}{exams} \\
\midrule
Passage \# & 319 & 313 & 369 & 399 & 167 & 168 & 273 & 244 \\
Question \# & 501 & 502 & 515 & 500 & 561 & 564 & 505 & 500 \\
Max Answer \# per Question & 3 & 3 & 3 & 3 & 1 & 1 & 1 & 1 \\
Max Evidence \# per Question & 2 & 2 & 3 & 3 & 2 & 2 & 4 & 4 \\
\midrule
Avg/Max Passage Tokens \# & 146/369 & 157/352 & 467/961 & 468/930 & 311/514 & 324/603 & 426/1096 & 413/1011  \\
Avg/Max Question Tokens \# & 12/28 & 11/28 & 15/37 & 15/37 & 15/39 & 16/55 & 14/28 & 14/31  \\
Avg/Max Answer Tokens \# & 3/25 & 3/27 & 6/64 & 5/33 & 6/20 & 6/27 & 7/25 & 7/35  \\
Avg/Max Evidence Tokens \# & 26/62 & 28/76 & 43/175 & 52/313 & 23/162 & 23/82 & 37/199 & 41/180  \\
\bottomrule
\end{tabular}
\caption{\label{dataset-stat} Statistics of the proposed ExpMRC.} 
\end{center}
\end{table*}

\section{ExpMRC}
\subsection{Subset Selection}
The motivation of our dataset is to provide a comprehensive MRC benchmark for evaluating not only the prediction accuracy but also how well it gives for its explanation.
To this end, our dataset is not completely composed of new data.
We adopt several well-designed MRC datasets as well as newly annotated data to form our dataset to minimize the repetitive annotations and place our work in line with previous works.

Specifically, our ExpMRC is partly developed from the following datasets, including two span-extraction MRC datasets and one multi-choice MRC dataset.
\begin{itemize}
	\item {\bf SQuAD} \citep{rajpurkar-etal-2016} is a well-known dataset for span-extraction MRC. Given a Wikipedia passage, the MRC system should extract a passage span as the answer to the question.
	\item {\bf CMRC 2018} \citep{cui-emnlp2019-cmrc2018} is also a span-extraction MRC dataset but in Chinese. Besides the traditional train/dev/test split, they also released a challenge set that requires multi-sentence inference while keeping the original span-extraction setting.
	\item {\bf C$^3$} \citep{sun-etal-2020-c3} is a Chinese multi-choice MRC dataset. The system should choose the correct option as the answer after reading the passage and question. To ensure the domain consistency for each subset, we only use non-dialogue subsets C$^3_\text{M}$.
\end{itemize}

As SQuAD and CMRC 2018 do not release the test set to the public, we are unable to adopt them directly.
Instead, we follow the original dataset construction steps to replicate two subsets for testing purposes, where the subsets are annotated from either English or Chinese Wikipedia passages.
Note that, during the subset annotation, we select the passages that do not appear in the original training and development set.

While we could have used RACE \citep{lai-etal-2017} as the counterpart of C$^3$, we decided not to adopt it. 
We had some in-house collected multi-choice MRC data, which is similar to RACE and is also designed for the middle and high school students.
More importantly, these data contain additional hints on the answering process, which are very helpful for the evidence annotation.
Thus, we decided to use our data instead of RACE.
We denote this new subset as RACE$^+$.

At this point, we have four subsets to be annotated, containing both span-extraction and multi-choice MRC in both English and Chinese.

\subsection{Evidence Annotation}
Before evidence annotation, the annotators are required to consider whether a question is appropriate for annotation.
We skipped some questions based on the following criteria.
\begin{itemize}
	\item The evidence span is a simple combination of the question and answer without much syntactical or semantical variance.
	\item The questions require external knowledge to be solved and cannot be only inferred from the passage.
	\item The conclusive questions of the whole passage, such as `what is the best title of this passage?', `what is the main idea of the passage?', etc.
\end{itemize}

After that, we begin the evidence annotation process.
First, the annotators are asked to read the question and the correct answer (passage span or option text).
Then, they should select (copy-and-paste) a passage span from the passage that can be the evidence of the answer. 
The evidence should be a minimal passage span that can support the answer and should not always be a complete sentence or clause. 
We encourage the annotators to select the evidence that needs reasoning skills, though this is not a usual case in these datasets, especially in span-extraction MRC (most of the questions do not need reasoning).

\subsection{Quality Assurance}
The annotators are paid about \$0.5 per evidence for all types of MRC data.
The annotators are either English-majored or Chinese-majored graduate students from China, depending on the language of the dataset.
Also, to avoid overworking and decreasing the annotation quality, we set a hard limit on the number of daily annotations for the evidence. 
After reaching the limit (300), the system is automatically locked and will be unlocked the next day. 

Besides, as our dataset has two or more referential evidence spans per question, we do not reveal the annotated evidence span to the current annotator to increase the diversity and avoid copy-and-paste behavior.
After the preliminary annotation, all evidence spans are checked one-by-one to ensure a high-quality dataset maximally. 
Finally, they should verify that we can pick out the correct answer by only reading the evidence and question to ensure that the annotation is valid.

\subsection{Statistics}
\begin{figure}[htp]
  \centering
  \includegraphics[width=1.0\columnwidth]{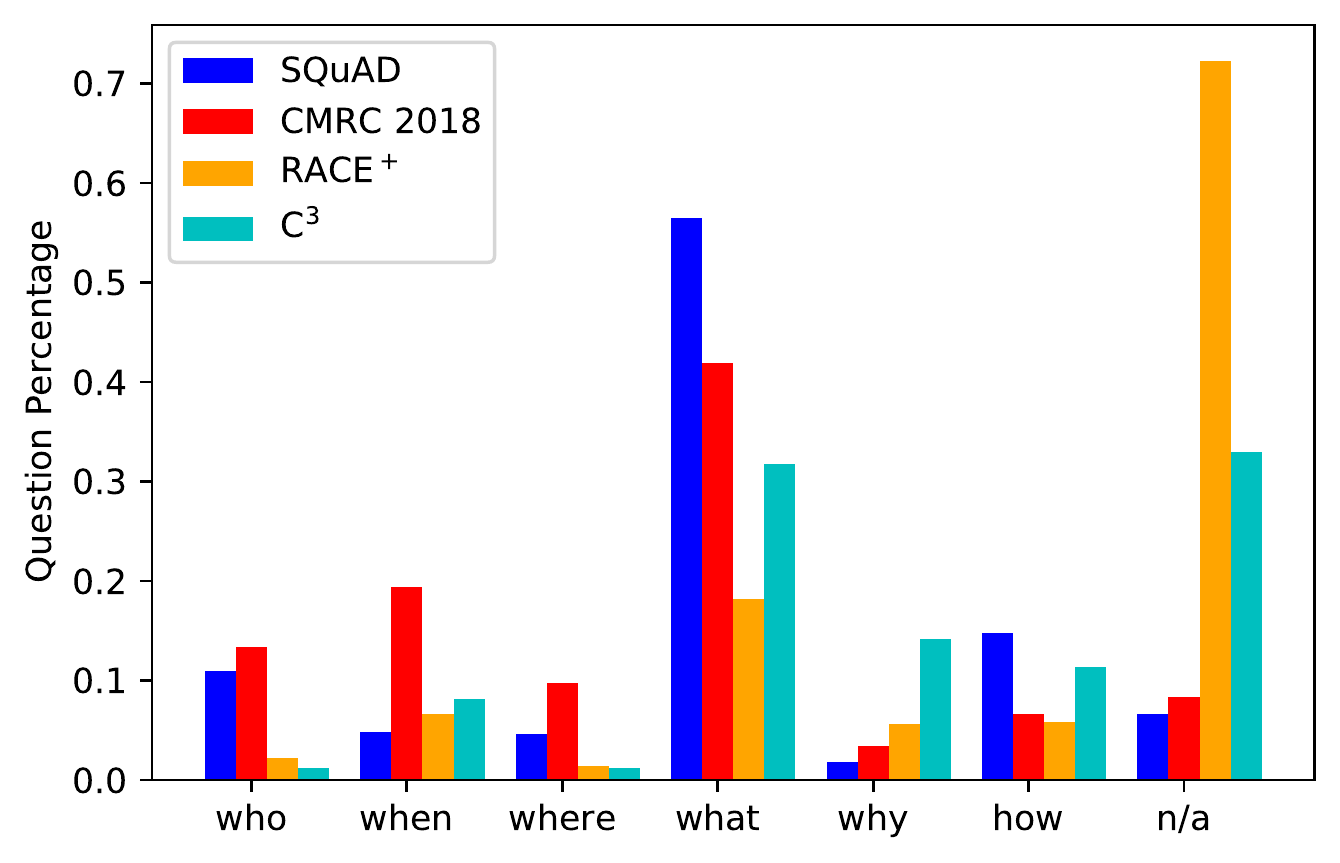}
  \caption{\label{question-type} Distribution of question types. } 
\end{figure}

The statistics of the proposed ExpMRC are listed in Table \ref{dataset-stat}.
Note that the `token' in Table \ref{dataset-stat} represents the character for Chinese and word for English.
For all subsets, we provide $2\sim4$ referential evidence spans for each question.
The distribution of the question type in each task's development set is depicted in Figure \ref{question-type}.
As we can see that there are fewer questions of {\em `who, when, where'} in RACE$^+$ and C$^3$, suggesting that these subsets are much more difficult.

\section{Baselines}
Given that the proposed ExpMRC is designed to evaluate the explainability in terms of the system's explanation text, we mainly focus on the {\em unsupervised approaches}\footnote{The term `unsupervised' specifically refers to the annotated explanation spans.} for our baseline systems.
We use pre-trained language models as the backbones to generate answers for the questions.
Then we apply several methods to generate evidence spans, where we classify them into non-learning and machine learning baselines.

\subsection{Non-learning Baselines}
For non-learning baselines, we mainly use the prediction and question as the clues for finding evidence.
For simplicity, we only consider extracting sentence-level evidence in these baselines, though the ground truth evidence may not always be a complete sentence.
We first split the passage into several sentences using `.!?' as delimiters.
Then we choose one of the passage sentences as the evidence prediction.
To find much accurate evidence sentence, we adopt three approaches.
\begin{itemize}
	\item {\bf Most Similar Sentence}: We calculate the token-level F1 score between the predicted answer span (or choice text) and each passage sentence. Then we select the sentence that has the highest F1 as the evidence prediction. In span-extraction MRC tasks, the extracted evidence should be the sentence that contains the prediction span in most of the cases.
	\item {\bf Most Similar Sentence with Question}: Similar to the `Most Similar Sentence' setting, but we use both question text and predicted answer span as the key to finding the most similar passage sentence.
	\item {\bf Answer Sentence}: Particularly, in span-extraction MRC tasks, we could directly extract the sentence that contains the answer prediction as the evidence.
\end{itemize}

These approaches largely rely on the accuracy of answer prediction, as the wrong prediction will directly affect the finding process of evidence.

\subsection{Machine Learning Baselines}
\begin{figure}[h]
  \centering
  \subfigure[Span-Extraction MRC]{\includegraphics[width=0.8\columnwidth]{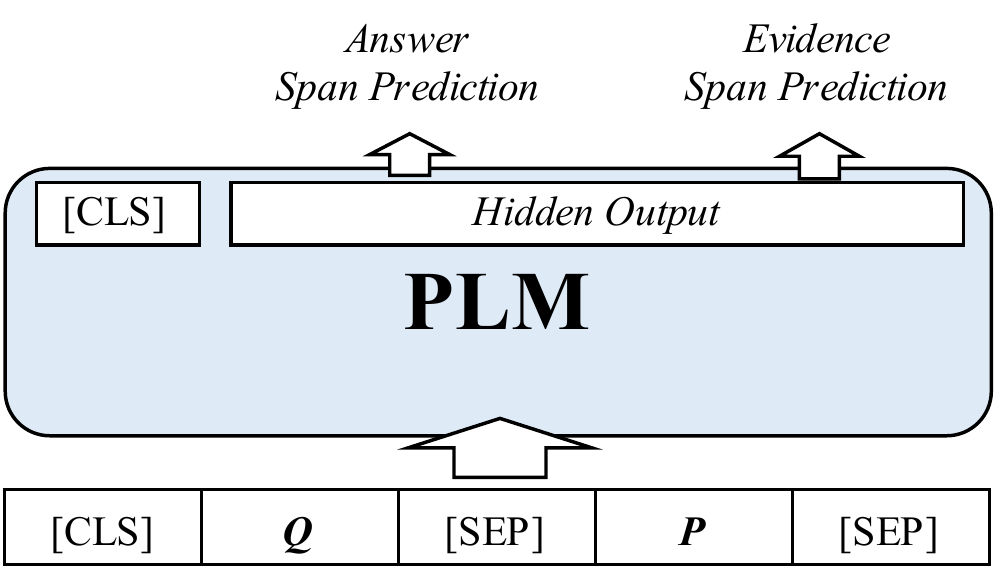}} \\
  \subfigure[Multi-Choice MRC]{\includegraphics[width=0.8\columnwidth]{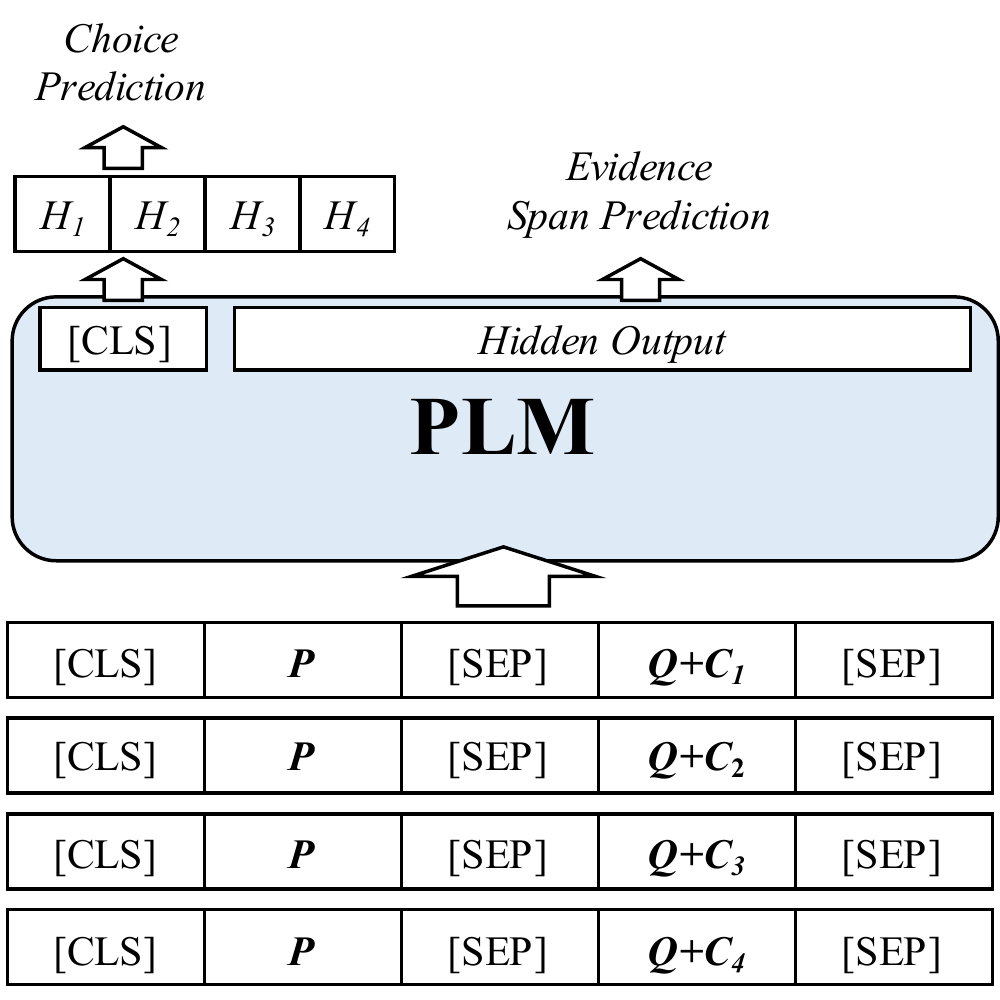}}
  \caption{\label{baseline-model} Neural network architecture of the baselines. } 
\end{figure}

As no training data is provided in ExpMRC, we seek a pseudo-training approach to accomplish a machine learning baseline system.
First, we generate pseudo-evidence for each sample in the respective training set, which has no evidence annotation.
We use the ground truth answer and question text to find the most similar passage sentence as the pseudo-evidence to form pseudo-training data.
Then we use the pseudo-training data and PLM to train a model that outputs both answer prediction and evidence prediction.
Specifically, we add an additional task head on top of the PLM's final hidden representation, alongside its original answer prediction task, as shown in Figure \ref{baseline-model}.
\begin{itemize}
	\item {\bf Span-Extraction MRC}: The concatenation of the question and passage is fed into PLM, and we use the final hidden representation with two fully-connected layers to predict the start and end positions of the answer span.
	\item {\bf Multi-Choice MRC}: The concatenation of the passage, question, and each choice is fed into PLM to get four pooled representations (assuming we have four candidate choices). Then we use a fully-connected layer with softmax activation to predict the final choice.
\end{itemize}

The evidence prediction is identical to the answer prediction in span-extraction MRC, where we project the final hidden representation $\bm{h} \in \mathbb{R}^{n \times h}$ into the start and end probability distributions $p^s, p^e \in \mathbb{R}^{n}$.
Then we calculate standard cross-entropy loss of the start and end positions for evidence span prediction.
\begin{gather}
    p^s = \mathbf{softmax}(\bm{h} \mathbf{w^s} + b^s) \\
    p^e = \mathbf{softmax}(\bm{h} \mathbf{w^e} + b^e) \\
	\mathcal{L}_{E} = -\frac{1}{2N} \sum\limits_{i=1}^{N}(y^{s}_i\log p^{s} + y^{e}_i\log p^{e})
\end{gather}

The final training loss is the sum of answer prediction loss $\mathcal{L}_A$ and the evidence prediction loss $\mathcal{L}_{E}$, where we apply $\lambda \in [0,1]$ scaling on $\mathcal{L}_{E}$, as the pseudo-training data is not quite accurate.
\begin{gather}
	\mathcal{L} = \mathcal{L}_{A} + \lambda \mathcal{L}_{E}
\end{gather}

\section{Evaluation Approach}
\subsection{Evaluation Metrics}
To evaluate how well the MRC model could generate explanations for the answers, we use the following metrics, which are divided into answer evaluation and evidence evaluation.

For answer evaluation, we strictly follow the original evaluation script for each subset.
Specifically, we use the F1-score (F1) to evaluate SQuAD and CMRC 2018. 
We dump Exact Match (EM) and only evaluated F1 for simplicity.
Note that, as these datasets are in different languages, the evaluation details are slightly different.
For RACE$^+$ and C$^3$, we use accuracy for evaluation.

For evidence evaluation, we also use F1 metrics, as most of the evidence spans are quite long, and it is hard for the machine to extract the evidence spans exactly, and thus we do not adopt EM.
Also, the central idea of the evidence is to provide enough information to support the answer, so it is proper to adopt F1.
Note that we only evaluate the correctness of evidence in this metric, regardless of the correctness of the answer.

Altogether, we also provide an overall F1 metric to give a comprehensive evaluation of the system.
For each instance, we calculate the score of the answer metric and evidence metric. The overall F1 of each instance is obtained by multiplying both terms.
Finally, the overall F1 of all instances is obtained by averaging all instance-level F1.
The overall F1 reflects the correctness of both answer and its evidence.
\begin{equation}
\mathtt{F1_{overall}} = \mathtt{F1_{answer}} \times \mathtt{F1_{evidence}}
\end{equation}

\subsection{Human Performance}
Following previous works \citep{rajpurkar-etal-2016,lai-etal-2017,cui-emnlp2019-cmrc2018}, we also report human performance to estimate how well does human performs on this dataset.
Following \citet{cui-emnlp2019-cmrc2018}, we use a {\em cross-validation approach} that regarding one of the candidates as prediction and treating the rest of the candidates as ground truths.
Final scores are obtained by averaging all possible combinations.
Specifically,
\begin{itemize}
	\item {\bf SQuAD, CMRC 2018}: In these datasets, both answer and evidence have multiple references, and thus we use the cross-validation approach for both and get their products as instance-level human performance. 
	\item {\bf RACE$^+$, C$^3$}: As these datasets have only one reference answer, we invite three annotators to answer a random set of 100 questions in each set to get averaged human performance of the answer. For the evidence, we directly use the cross-validation approach for the selected random set. Similarly, the instance-level human performance is obtained by the product of the answer and evidence score.
\end{itemize}

Note that as the evidence spans are annotated by referring to either the answers or additional hints, the actual human performance could be much lower, and thus these results should be regarded as {\em ceiling} human performance roughly.
Finally, we average the scores in all instances to obtain the final overall human performance.

\begin{table*}[htp]
\small
\begin{minipage}{\textwidth}
\begin{tabular}{l c c c c c c c c c c c c}
\toprule
\multirow{2}*{\bf System} & \multicolumn{3}{c}{\bf SQuAD (dev)} & \multicolumn{3}{c}{\bf SQuAD (test)}& \multicolumn{3}{c}{\bf CMRC 2018 (dev)} & \multicolumn{3}{c}{\bf CMRC 2018 (test)} \\
& \bf Ans. & \bf Evi. & \bf All & \bf Ans. & \bf Evi. & \bf All & \bf Ans. & \bf Evi. & \bf All & \bf Ans. & \bf Evi. & \bf All \\
\midrule
\em Human Performance & \em 90.8 & \em 92.1 & \em 83.6 & \em 91.3 & \em 92.9 & \em 84.7 & \em 97.7 & \em 94.6 & \em 92.4 & \em 97.9 & \em 94.6 & \em 92.6 \\
\midrule
\multicolumn{8}{l}{\em PLM base-level Baselines} \\
Most Similar Sent. & 87.4 & 81.8 & 74.5
& 87.1 & 85.4 & 76.1
& 82.3 & 71.9 & 60.1 
& 84.4 & 62.2 & 52.9 \\
Most Similar Sent. w/ Ques. & 87.4 & 81.0 & 72.9
& 87.1 & 84.8 & 75.6
& 82.3 & 76.9 & 63.9 
& 84.4 & \bf 69.8 & \bf 59.9 \\
Predicted Answer Sent. & 87.4 & \bf 84.1 & \bf 76.4
& 87.1 & \bf 89.1 & \bf 79.6 
& 82.3 & \bf 78.0 & \bf 66.8 
& 84.4 & 69.1 & 59.8 \\
Pseudo-data Training & \bf 88.1 & 76.9 & 69.7 
& \bf 88.1 & 76.3 & 67.5 
& \bf 82.7 & 65.1 & 54.7 
& \bf 85.7 & 59.7 & 51.2 \\
\midrule
\multicolumn{8}{l}{\em PLM large-level Baselines} \\
Most Similar Sent. & 93.0 & 83.9 & 79.3
& 92.3 & 85.7 & 80.4
& 82.8 & 71.6 & 60.3 
& 88.6 & 63.0 & 55.9 \\
Most Similar Sent. w/ Ques. & 93.0 & 81.9 & 77.4
& 92.3 & 85.1 & 79.8
& 82.8 & 76.3 & 63.6 
& 88.6 & \bf 71.0 & 63.2 \\
Predicted Answer Sent. & 93.0 & \bf 85.4 & \bf 81.8
& 92.3 & \bf 89.6 & \bf 83.6
& 82.8 & \bf 77.7 & \bf 66.9 
& 88.6 & 70.6 & \bf 63.3 \\
Pseudo-data training & \bf 93.3 & 78.1 & 73.5
& \bf 93.4 & 76.1 & 72.1 
& \bf 83.8 & 65.3 & 55.6 
& \bf 88.8 & 58.4 & 51.5 \\
\bottomrule
\end{tabular}
\end{minipage}
\\ \\ \\
\begin{minipage}{\textwidth}
\begin{tabular}{l c c c c c c c c c c c c}
\toprule
\multirow{2}{*}{\bf System} & \multicolumn{3}{c}{\bf RACE$^+$ (dev)} & \multicolumn{3}{c}{\bf RACE$^+$ (test)}& \multicolumn{3}{c}{\bf C$^3$ (dev)} & \multicolumn{3}{c}{\bf C$^3$ (test)} \\
& \bf Ans. & \bf Evi. & \bf All & \bf Ans. & \bf Evi. & \bf All & \bf Ans. & \bf Evi. & \bf All & \bf Ans. & \bf Evi. & \bf All \\
\midrule
\em Human Performance & \em 92.0 & \em 92.4 & \em 85.4 & \em 93.6 & \em 90.5 & \em 84.4 & \em 95.3 & \em 95.7 & \em 91.1 & \em 94.3 & \em 97.7 & \em 90.0 \\
\midrule
\multicolumn{8}{l}{\em PLM base-level Baselines} \\
Most Similar Sent. & 62.4 & 36.6 &28.2
& 59.8 & 34.4 & 26.3
& 68.7 & 57.7 & \bf 47.7 
& 66.8 & 52.2 & 41.2 \\
Most Similar Sent. w/ Ques. & 62.4 & 44.5 & 31.5
& 59.8 & 41.8 & \bf 27.3
& 68.7 & \bf 62.3 & 47.3 
& 66.8 & 57.4 & \bf 42.3 \\
Pseudo-data training & \bf 63.6 & \bf 45.7 & \bf 31.7
& \bf 60.1 & \bf 43.5 & 27.1
& \bf 70.9 & 59.9 & 43.5 
& \bf 69.0 & \bf 57.5 & 40.6 \\
\midrule
\multicolumn{8}{l}{\em PLM large-level Baselines} \\
Most Similar Sent. & \bf 69.0 & 37.6 & 29.9
& 68.1 & 36.8 & 28.9 
& 73.1 & 59.4 & 49.9 
& 72.0 & 52.7 & 43.9 \\
Most Similar Sent. w/ Ques. & \bf 69.0 & \bf 48.0 & \bf 36.8
& 68.1 & \bf 42.5 & \bf 31.3
& 73.1 & 63.2 & \bf 50.9 
& 72.0 & 58.4 & 46.0 \\
Pseudo-data training & \bf 69.0 & 45.9 & 32.6 
& \bf 70.4 & 41.3 & 30.8
& \bf 76.4 & \bf 64.3 & 50.7 
& \bf 74.4 & \bf 59.9 & \bf 47.3 \\
\bottomrule
\end{tabular}
\end{minipage}
\caption{\label{result-baseline} Baseline results on SQuAD, CMRC 2018, RACE$^+$, and C$^3$. } 
\end{table*}

\section{Experiments}
\subsection{Setups}
We use pre-trained language models as backbones of baseline systems.
Specifically, we use BERT-base and BERT-large-wwm \citep{devlin-etal-2019-bert} for English, and MacBERT-base/large \citep{cui-etal-2020-revisiting} for Chinese tasks.
We use a universal initial learning rate of 3e-5 and iterate two training epochs for all tasks.
The maximum sequence length is set to 512, and a QA length of 128 in all experiments.
We use \textsc{Adam} \cite{kingma2014adam} with weight decay optimizer for training.
We set $\lambda=0.1$ in the final loss function to penalize the evidence pseudo-data training, which we found to be effective, and further investigation is discussed in Section \ref{answer-evidence-balance}.

\subsection{Baseline Results}
Overall, the best-performing baselines are still far behind the human performance, indicating that the proposed dataset is challenging.
Also, the gaps in multi-choice MRC subsets are bigger than the ones in span-extraction MRC.
For all subsets, adding question text for similarity calculation is more effective than only using the predicted answer.
For span-extraction MRC, traditional token similarity methods seem to be more effective as the answer is already a passage span, and its evidence often lies around its context.
On the contrary, the pseudo-data training approach is more effective in multi-choice MRC, where the options are not composed by the passage span, which is not capable of direct mapping, and it requires similarity calculation in semantics but not only in the token-level calculation.

\begin{figure*}[tp]
  \centering
  \subfigure[SQuAD]{\includegraphics[width=0.51\columnwidth]{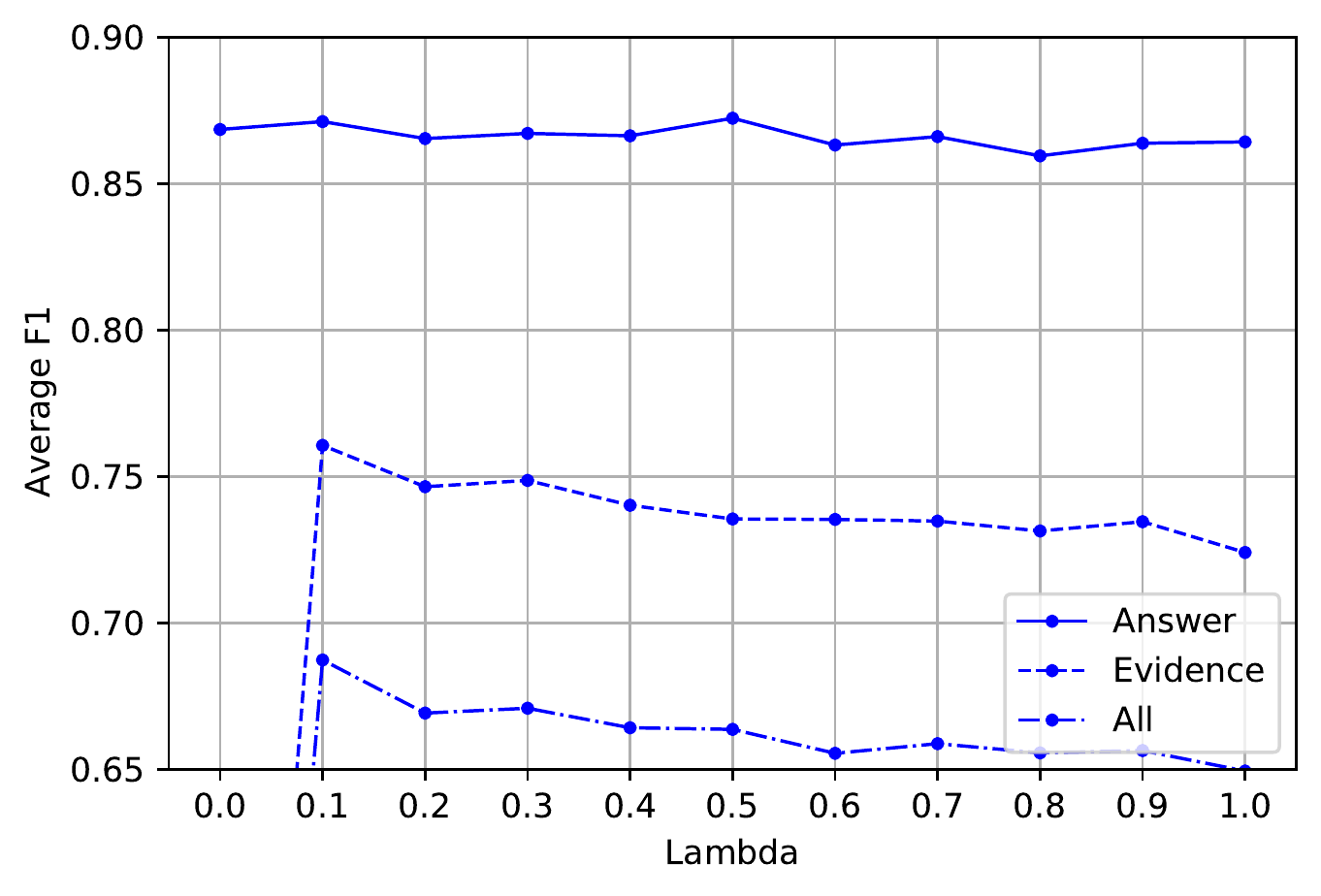}} 
  \subfigure[CMRC 2018]{\includegraphics[width=0.51\columnwidth]{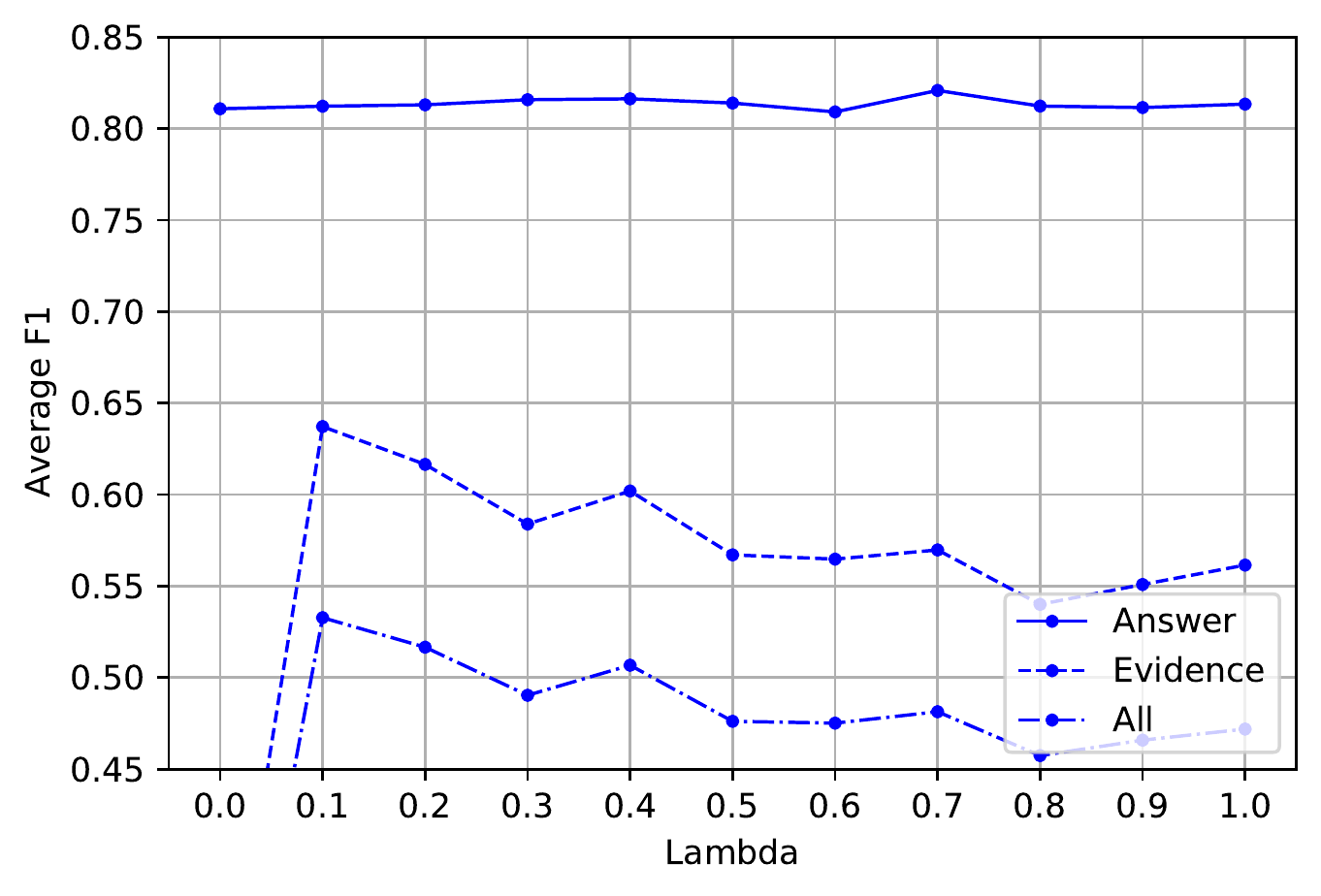}}
  \subfigure[RACE$^+$]{\includegraphics[width=0.51\columnwidth]{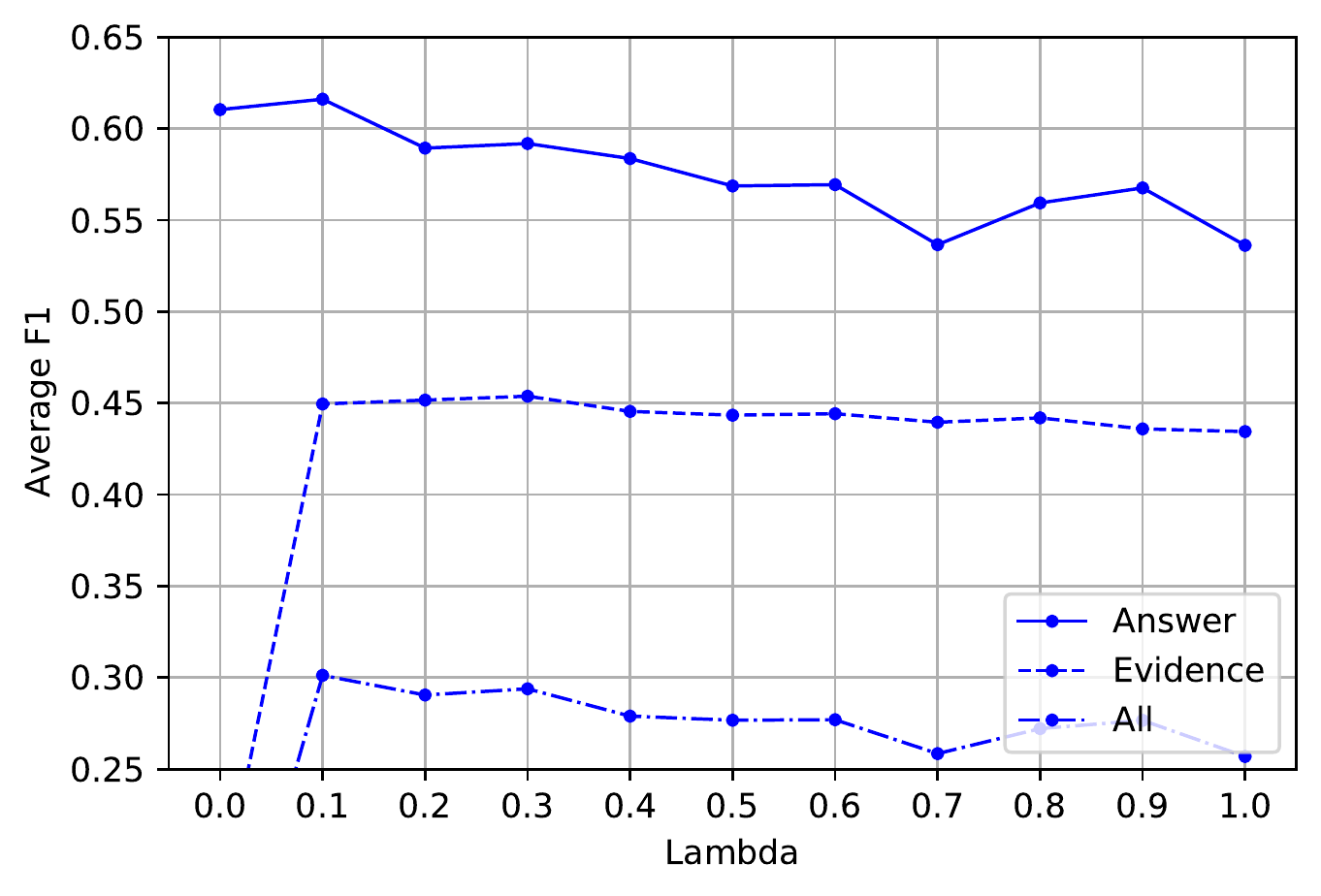}}
  \subfigure[C$^3$]{\includegraphics[width=0.51\columnwidth]{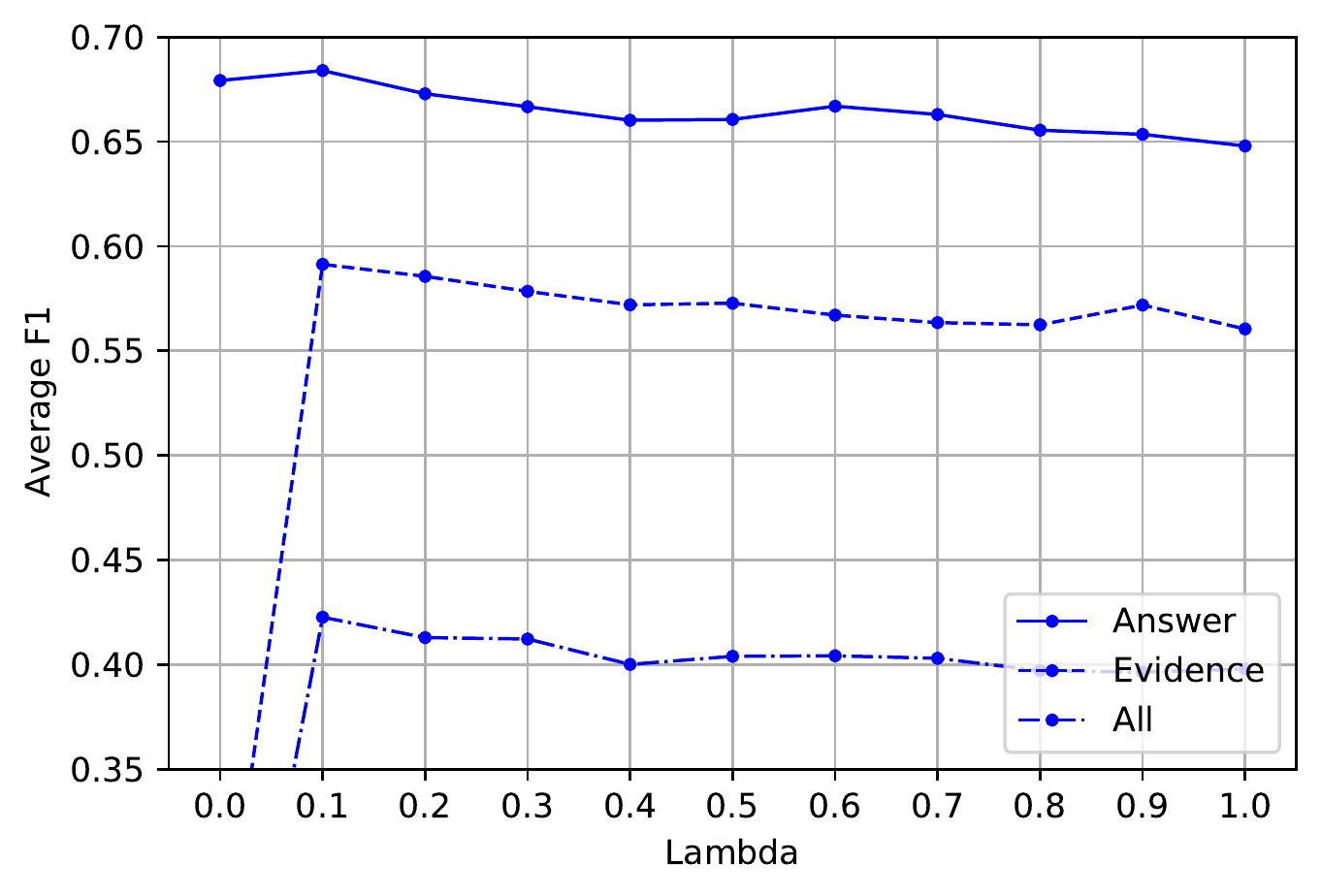}}
  \caption{\label{analysis-lambda} Effect of the lambda term in the evidence loss for different tasks. } 
\end{figure*}

The improvement of both answer and evidence prediction does NOT necessarily improve the overall score. 
For example, in the development set of C$^3$, pseudo-data training large-level baseline yields better performance on both answer and evidence prediction than the others.
However, its overall score of 50.7 is lower than the best-performing baseline of 50.9.
After checking the prediction file, we discovered that there are more samples that have either better evidence spans for the wrong answer prediction or worse evidence spans for correct answer prediction, which decreases the overall score.

Another interesting observation is that though pseudo-data training baselines do not yield better overall scores mostly, we have seen almost consistent improvements in the answer prediction accuracy, such as in C$^3$ using large-level PLM (e.g., dev +3.3, test +2.4). 
This suggests that using pseudo evidence helps improve answer prediction, and we expect there will be another improvement when we use a more efficient way to extract high-quality pseudo evidence.

\subsection{Answer and Evidence Balance}\label{answer-evidence-balance}
For balancing the ratio between the answer and evidence loss, we applied a lambda term on the evidence loss.
To explore the effect of the lambda term, we select different $\lambda \in [0,1]$, and plot the 5-run average performance of each task.
The results are shown in Figure \ref{analysis-lambda}.

Overall, as we can see that by increasing the lambda term, the evidence score and overall score decrease, suggesting that the pseudo-data training could not be regarded as important as the original supervised task training (answer prediction), as the pseudo-data is not constructed by the golden evidence.
However, when it comes to the answer score, we observe that the span-extraction MRC tasks are less sensitive to the lambda term compared to the multi-choice MRC tasks. 

\subsection{Upper Bound Test}
In the last section, we analyze the possible next steps to achieve better evidence extraction performance.
Besides the `Most Similar Sentence with Question' (MSS w/ Ques.) and `Predicted Answer Sentence' (PA Sent.), we also provide two additional baselines.
We extract the sentence that contains the ground truth answer (GA Sent.) or annotated evidence (GE Sent.) to measure the upper bounds for those systems that only extract sentence-level evidence.
The results are shown in Table \ref{evidence-analysis}.

\begin{table}[htp]
\small
\begin{center}
\begin{tabular}{l cccc}
\toprule
& \bf SQuAD & \bf CMRC & \bf RACE$^+$ & \bf C$^3$ \\
\midrule
MSS w/ Ques. & 81.9 & 76.3 & 48.0 & 63.2 \\
PA Sent. & 85.4 & 77.7 & - & - \\
GA Sent. & 88.2 & 82.1 & 49.9 & 66.8 \\
GE Sent. & 91.6 & 85.2 & 86.9 & 89.1 \\
\em Human Perf. & \em 92.1 & \em 94.6 & \em 92.4 & \em 95.7 \\
\bottomrule
\end{tabular}
\caption{\label{evidence-analysis} Evidence performance of development sets.} 
\end{center}
\end{table}

As we can see, the PA-GA and GA-GE gap in span-extraction MRC is very small (about 3\%$\sim$5\%), suggesting that the current system is about to reach the ceiling performance when only using sentence-level evidence extraction.
On the contrary, in multi-choice MRC, we see a huge gap between GA and GE, indicating that only using the answer sentence is not enough to achieve strong evidence extraction performance. 

The gap between GE and human performance indicates the gains from expanding sentence-level evidence to a free-form evidence span.
Besides the SQuAD task, the others yield 5.5\%$\sim$9.4\% gap, which demonstrates that finding the exact evidence span in these tasks could still achieve a decent improvement.

\section{Conclusion}
In this paper, we propose a comprehensive benchmark for evaluating the explainability of machine reading comprehension systems.
The proposed ExpMRC benchmark contains four datasets, covering span-extraction MRC and multiple-choice MRC in both English and Chinese. 
ExpMRC aims to evaluate the MRC system to give not only correct predictions on the final answer but also extract correct evidence for the answer.
We set up several baseline systems to thoroughly evaluate the difficulties of ExpMRC.
The experimental results show that both traditional and state-of-the-art pre-trained language models still underperform human performance by a large margin, indicating that more work should be done on explanation extraction.
We hope the release of the dataset could further accelerate the research of explainability and interpretability for the MRC systems.

\bibliography{acl2021}
\bibliographystyle{acl_natbib}

\end{CJK*}

\end{document}